# Detecting Figures and Part Labels in Patents: Competition-Based Development of Image Processing Algorithms

Christoph Riedl,* Richard Zanibbi,* Marti A. Hearst, Siyu Zhu, Michael Menietti, Jason Crusan, Ivan Metelsky, and Karim R. Lakhani

***Abstract*—We report the findings of a month-long online competition in which participants developed algorithms for augmenting the digital version of patent documents published by the United States Patent and Trademark Office (USPTO). The goal was to detect figures and part labels in U.S. patent drawing pages. The challenge drew 232 teams of two, of which 70 teams (30%) submitted solutions. Collectively, teams submitted 1,797 solutions that were compiled on the competition servers. Participants reported spending an average of 63 hours developing their solutions, resulting in a total of 5,591 hours of development time. A manually labeled dataset of 306 patents was used for training, online system tests, and evaluation. The design and performance of the top-5 systems are presented, along with a system developed after the competition which illustrates that winning teams produced near state-of-the-art results under strict time and computation constraints. For the 1st place system, the harmonic mean of recall and precision (f-measure) was 88.57% for figure region detection, 78.81% for figure regions with correctly recognized figure titles, and 70.98% for part label detection and character recognition. Data and software from the competition are available through the online UCI Machine Learning repository to inspire follow-on work by the image processing community.**

## I. Introduction

The United States Patent and Trademark Office (USPTO) is in the process of bringing an archive of eight million patents into the digital age by modernizing the representation of these documents in its information technology systems. In their daily work, patent examiners at the USPTO, as well as patent lawyers and inventors throughout the world, rely on this patent archive. Locating existing patents related to new patent application requires significant effort, which has motivated research into automatic retrieval of patents using both text [1] and images [2]. Most USPTO patent documents contain drawing pages which describe the invention graphically. By convention and by rule, these drawings contain figures and parts that are annotated with numbered labels but not with text, and so readers must scan the entire document to find the meaning of a given part label.

One would like to be able to automatically link part labels with their definitions in digital patent documents to save readers this effort. For example, one could create 'tool-tips' for part labels and figures, where hovering the pointer over a part label or figure brings up text describing the part or figure, reducing the need to switch back and forth between diagram and text pages. Unfortunately, robust solutions to this problem are currently unavailable. While document image analysis [3] and optical character recognition [4] have made significant advances, detecting figures and labels scattered within drawings remains a hard problem. More generally, text detection in documents and natural scenes [5], [6], [7], [8] remains a challenging image processing task.

Prize-based competitions have a long history of encouraging innovation and attracting unconventional individuals who can overcome difficult challenges and successfully bridge knowledge domains [9], [10]. This has recently lead to an emergence of commercial platforms including TopCoder, InnoCentive, and Kaggle that have specialized in supporting and executing large-scale competitions around algorithm and software development. In September 2009, President Obama called on all U.S. Federal government agencies to increase their use of prizes and competitions to address difficult challenges. Following this, the U.S. Congress granted all those agencies authority to conduct prize competitions to spur innovation in the America COMPETES Reauthorization Act of 2010 [11]. These developments helped provide a legal path for government agencies to conduct prize competitions. NASA, which already had prize authority and deep experience working with the TopCoder software competition community [12], opened a Center of Excellence for Collaborative Innovation to help other U.S. Federal Agencies run challenges.

These developments together led to the USPTO launching a software challenge to develop image processing algorithms

*Corresponding authors.

C. Riedl is with D'Amore-McKim School of Business, and College of Computer & Information Science, Northeastern University, Boston, MA 02115 (e-mail: c.riedl@neu.edu).

R. Zanibbi is with Department of Computer Science, Rochester Institute of Technology, Rochester, NY 14623 (e-mail: rlaz@cs.rit.edu).

M. A. Hearst is with School of Information, UC Berkeley, Berkeley, CA 94720 (e-mail: hearst@berkeley.edu).

S. Zhu is with Center for Imaging Science Rochester Institute of Technology, Rochester, NY 14623 (e-mail: sxz8564@rit.edu).

M. Menietti is with Institute for Quantitative Social Science, Harvard University, Cambridge, MA 02138 (e-mail: mmenietti@fas.harvard.edu).

J. Crusan is with Advanced Exploration Systems Division at NASA, Washington DC (e-mail: jason.crusan@nasa.gov).

I. Metelsky is with TopCoder Inc., Glastonbury, CT 06033 (e-mail: imetelsky@topcoder.com).

K. R. Lakhani is with Department of Technology and Operations Management, Harvard Business School, Boston, MA 02134 (e-mail: klakhani@hbs.edu).

CR, MM, KJB, and KRL are also with Harvard-NASA Tournament Laboratory, Harvard University, Cambridge, MA 02138.

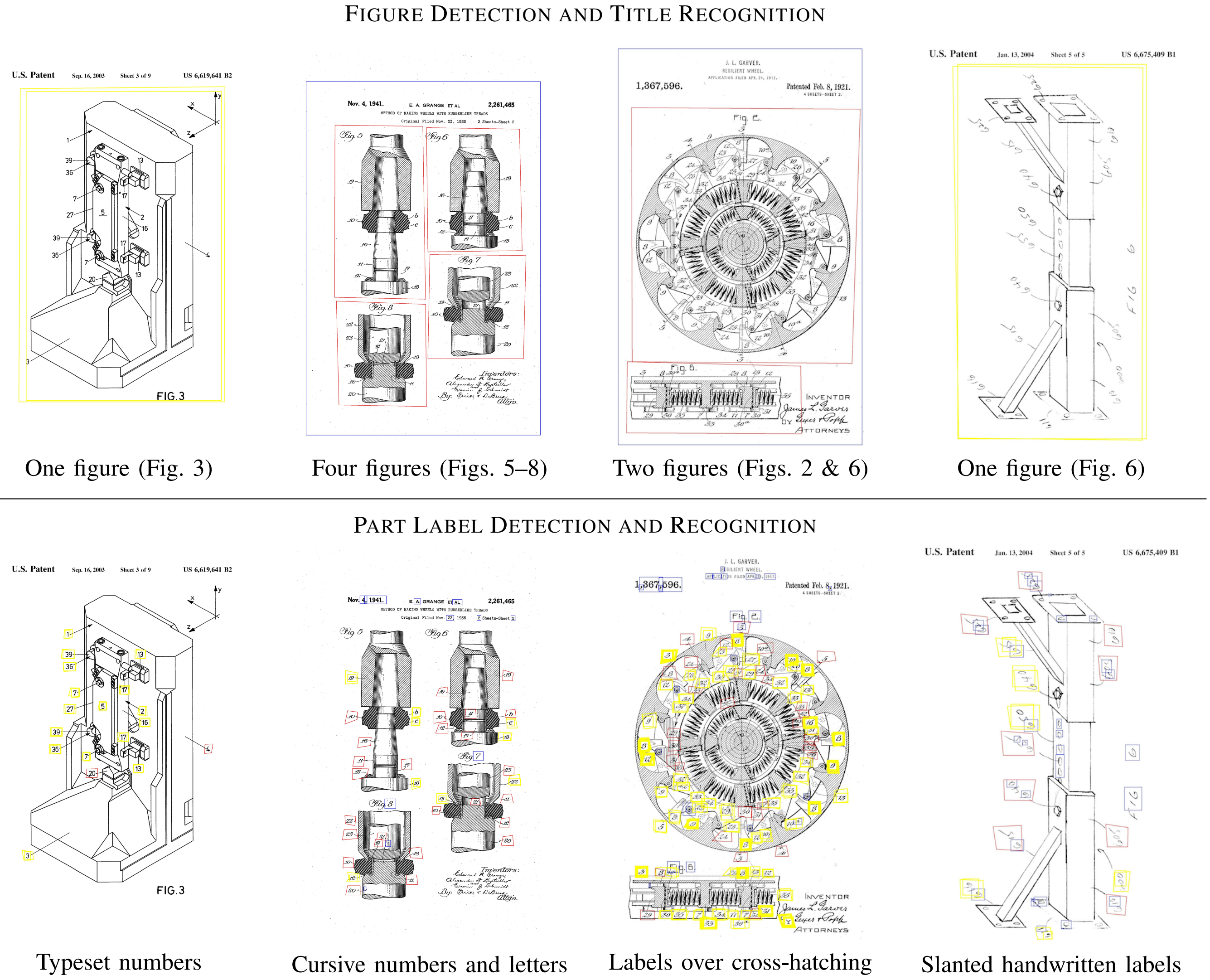

Fig. 1. Sample results from the 1st place system (leftmost two columns) and 2nd place system (rightmost two columns) on the Figure Detection and Title Recognition task (top row) and the Part Label Detection and Text Recognition Task (bottom row). Target regions are shown in yellow (located, or *true positive*), blue (*false positive*), and red (missed, or *false negative*). Figure titles are correctly recognized when they are included in the figure region box and the text label for the figure region box contains the correct title. On the top row, inner columns show multi target figures (red) which the algorithm mistakenly merges into a single figure (outer blue rectangles). Figure regions and titles are recognized correctly in the outer examples.
The bottom row shows part label detection and recognition results. In the leftmost figure, all but one label is detected by the 1st place algorithm, but in the second column from left, it misses several labels because of the font used. The example in the third column has dense line art and hatching. The 2nd-place algorithm's character recognizer can better handle the cursive font used, but a number of labels are missed (red) including labels touching or on top of lines. In the rightmost example, the page is rotated (in landscape orientation). Here six labels are detected correctly by the 2nd-place algorithm, but a number of false positives (blue) are caused by over-segmenting part labels. Additional false positives (blue) are produced for holes in the diagram (which have the same shape as 0/O), and the figure title (at right).

to recognize figure and part labels in patent documents on the TopCoder platform in December 2011 [13]. The goal of the competition was to detect figure locations and labels along with part labels in patent drawings, to enable their use in cross-referencing text and image data.

The specific goal of the challenge was to extract the following from patent drawing pages: 1) figure locations and titles, and 2) part label locations and text. Each region type was represented by a rectangle (bounding box) with a text label (e.g., as a triple $((20, 20), (100, 100), \text{“}9b\text{”})$ representing part label ‘9b’ located in a rectangle with top-left corner $(20, 20)$ and bottom-right corner $(100, 100)$). Inputs and outputs for competition systems are provided in Table I.

TABLE I
USPTO CHALLENGE OVERVIEW

| |
|---|
| INPUTS |
| 1. 8-bit greyscale drawing page scan (72 dpi) |
| 2. Associated HTML patent text (*Claims* and *Description* sections) |
| OUTPUTS |
| 1. Figure bounding boxes and title text |
| 2. Part label bounding boxes and text |

Participants were provided with images of patent drawing pages, each of which contains one or more figures (see Fig. 1). Each figure has a title, and in most cases a large number of part numbers affixed to their parts with curved lines and arrows. Most part labels are numerical or alphanumerical. Complicating matters, many drawing pages also include additional numbers and text, such as page numbers, dates, patent numbers, and inventor names.

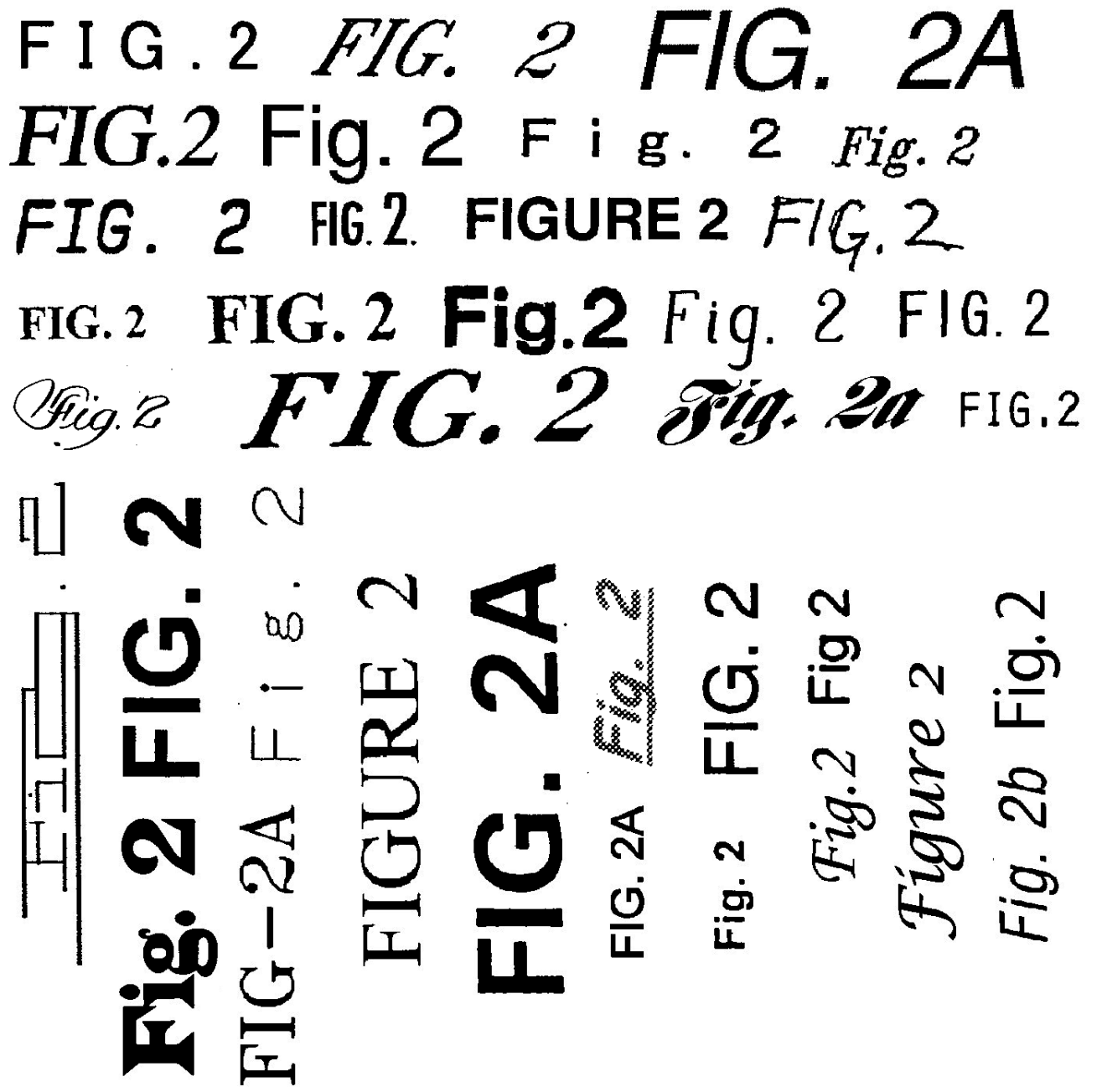

Fig. 2. Figure Titles from the USPTO Competition Data Set

Each drawing page image is accompanied by the text of the associated patent in HTML format. These are useful because figures are described explicitly in most recent patents, and part labels must be referred to at least once in the text. Participants could use the HTML text to validate and modify character recognition output.

Fig. 1 illustrates results from the 1st place system (leftmost two columns) and 2nd place system (rightmost two columns). Both target ('ground truth') regions and regions detected by the algorithms are shown. The drawing page in the leftmost column contains one figure titled 'FIG. 3' which has 18 part labels. A number of challenges are shown in Fig. 1 including differing page orientations (portrait vs. landscape), multiple figures on a page, text that does not belong to figures or part labels, different fonts and font styles, handwritten text, rotated text, drawing elements that look like characters, and part labels that intersect lines in a drawing.

A sample of figure titles from the USPTO competition data are shown in Fig. 2. There is a large variety of fonts, font styles (bold, italic, underline), and formats (e.g., 'Fig. 2,' 'Fig 2,' 'FIG-2,' and 'Figure 2'), in both portrait and landscape orientation. For the most part, part labels are typeset numeric ('11') or alphanumeric ('14b') strings in either portrait or landscape orientation. As illustrated in Fig. 1, there are also a number of drawing pages for which part labels are handwritten at an angle (i.e., slanted).

In this paper, we present the protocol and results of this competition. Section II describes related work. Section III describes the challenge in detail. In Section IV we describe the approaches used in the top-5 algorithms submitted to the competition and analyze their performance in Section V. The top-5 ranked systems adopted similar strategies, but differed in their approaches to text/graphics separation, page orientation detection, region segmentation, character recognition (OCR), and validation.

In Section VI we discuss the implications of this work. We provide details on the competition design which could serve as a template for other competitions that aim at solving image processing problems by drawing on a global pool of talent [14]. The diversity of approaches resulting from competitions has the potential to create new knowledge in image processing where the computational or algorithmic problem may represent a barrier to rapid progress but where finding the solution is not itself the major thrust of current scientific efforts.

Software for the top-5 systems along with the complete labeled training and test data have been released as open source under the Apache License 2.0. By doing this, we hope to encourage other researchers to pick up the work we have done and develop it further. The data and source code are available from the UCI Machine Learning Repository.[1]

## II. Related Work

In this section, we present background on competition-based algorithm and software development in general, and competitions in document image analysis and information (patent) retrieval in particular. We then provide an overview of document image analysis and graphics recognition, which provides a foundation for the analysis of the top-5 systems later in the paper, and finally a discussion of work in text and dimension recognition, to provide context for analyzing the algorithms developed for the competition.

Prize-based competitions have driven innovation throughout history [15]. For example, in the 18th century the British government announced a prize of £20,000 for finding a method to determine the longitude of a ship's location. More recently, prize-based competitions have been used to find solutions to hard algorithmic problems in biotech and medical imaging [9], [16]. These competitions provide an alternative to approaches requiring an extensive search to identify and contract with potential solvers.

In recent years, prize-based contests have emerged as part of a major trend towards solving industrial R&D, engineering, software development, and scientific problems. In the popular press, such competitions are often referred to as 'crowdsourcing' [17]. In general, crowdsourcing has come to imply a strategy that relies on external, unaffiliated actors to solve a defined problem [9]. Competitions provide an opportunity to expose a problem to a diverse group of individuals with varied skills, experience, and perspectives [18], some of whom may be intrinsically motivated, e.g., by the desire to learn or gain reputation within a community of peers. They also allow rapid exploration of multiple solutions in parallel by participants.

[1] http://archive.ics.uci.edu/ml/datasets/USPTO+Algorithm+Challenge%2C+run+by+NASA-Harvard+Tournament+Lab+and+TopCoder++++Problem%3A+Pat

Within the engineering and computer science communities, competitions at academic conferences are common. Specifically for document image analysis, the International Conference on Document Analysis and Recognition (ICDAR [19]), the International Association for Pattern Recognition (IAPR) International Workshop on Graphics Recognition (GREC [20]), and the International Workshop on Document Analysis Systems (DAS [21]) have hosted numerous competitions for a variety of document image analysis tasks.

More broadly, some of the best-known and most highly regarded academic competitions held within computer science are the Text REtrieval Conference (TREC) competitions, held for over two decades to develop and refine algorithms for text and multimedia-based searches [22]. In the past, TREC has included a competition for patent retrieval [23]. In recent years, the Cross-Language Evaluation Forum (CLEF) has held competitions on patent retrieval (including drawing images within queries), and on recognition of common patent diagrams such as chemical diagrams and flow charts [1].

While tremendously valuable for discerning and advancing the state-of-the-art, participants in academic competitions tend to belong to the community associated with a particular conference, prize amounts (if any) are small, and often a conference participation fee is required. In the USPTO image processing competition described in this article, crowdsourcing with cash prizes for top-placing systems was used to solicit solutions from a global pool of talent reaching beyond the academic image processing community.

### *A. Document Image Analysis and Graphics Recognition*

In the USPTO Challenge competition, the goal was to determine the location and contents of figures and part labels in U.S. patent drawing page images, with access to patent text provided as an HTML document. We provide a brief review of Document Image Analysis (DIA) and graphics recognition in this section, followed by a more detailed summary of work closely related to the competition in the next section.

Document image analysis and optical character recognition (OCR) were among the earliest applications of Artificial Intelligence [3]. An overview of current practice will appear in an upcoming handbook [4]. There have been decades of work on detecting and recognizing text in pages that are noisy, skewed, written in multiple languages, and that include graphics of various sorts (e.g., math, tables, flowcharts, engineering diagrams), along with the recognition of graphics themselves. Systems have scaled up to the point where patterns and statistics inferred from the contents of entire books are used to improve accuracy [24].

DIA systems must address the fundamental pattern recognition problems. These include identifying regions of interest (segmentation), determining region types (classification), and how regions are related/structured (parsing). These problems interact in complex ways. As an example, Casey and Lecolinet's well-known survey identifies three main approaches to character segmentation: dissection (cutting the images using image features), recognition-based (incorporating OCR output into available features), and holistic (classifying whole words rather than characters [25]). Most sophisticated character segmentation methods are recognition-based, with the final segmentation maximizing a criterion based on probabilities (e.g., from a Hidden Markov Model) or costs associated with recognized characters.

A key challenge is to generate as many valid hypotheses as possible (i.e., maximizing *recall*), while limiting the number of invalid hypotheses produced (i.e., maximizing the *precision* of *generated* hypotheses). A trade-off is encountered here: To obtain high recall in the presence of noise (e.g., for the USPTO competition, from touching characters in a figure title, or part labels intersecting lines in a drawing), a recognition system needs to generate alternative hypotheses. However, as more hypotheses are generated, execution time increases and the likelihood of missing valid hypotheses and accepting invalid hypotheses increases. To quantify this aspect of hypothesis generation, recall and precision metrics measured over generated hypotheses have been proposed [26], along with tools to make this information easier to collect and analyze [27].

In general, systems employing language models for specific documents perform best [3]. A concise, well-fit language model provides numerous beneficial constraints on system behavior, supporting efficient generation and accurate selection of hypotheses. As a simple example, recognizing a word comprised of an arbitrary number of Latin characters (e.g., a-z, A-Z) is much more difficult than recognizing the text of a U.S. postal code, which has a fixed length (5 characters), contains only digits (0-9), and for which a largely complete set of valid codes is available (i.e., an explicit language model, given by an enumeration of known valid codes). In this case, invalid codes may be detected with high reliability and can be replaced by similar, yet valid, codes.

State-of-the-art systems process images through bottom-up (from the image) and top-down (from the language model) methods [4]. Commonly, hypotheses are generated bottom-up from image features, and then ranked, filtered, and revised using syntactic and probabilistic constraints in a post-processing stage. For example, confusing ‘i’ for ‘1’ in the word ‘sin’ may be corrected using character bi-gram dictionaries or probabilities [28]. This ‘generate-and-test/revise’ approach is also frequently applied to problems at higher levels of abstraction, such as determining the type and spatial arrangement of page regions (e.g., for text and graphics regions using Markov Random Fields [29]).

Alternatively, hypotheses may be generated and revised incrementally as a system tries to parse a page, generating hypotheses top-down from a language model and then validating them bottom up, or vice versa. Researchers have used probabilistic language models for text in document pages (e.g., based on Hidden Markov Models [30] or Stochastic Context-Free Grammars [31]) and graphics such as mathematical notation [32]. Context-free attribute grammar models [33] have been applied to recognizing engineering drawings [34], tables [35] and other graphic types.

Recognizing engineering and architectural drawings is closely related to the USPTO challenge, in the sense that detecting objects is similar to figure detection, and recognizing dimensions is similar to recognizing part labels in patents. As an example of a syntax-driven system, Lu et al. consulted an expert to help design their knowledge-based system for architectural drawing recognition [34], and observed that many implicit relationships need to be interpreted, such as symmetry markings indicating inherited properties of objects, and the extensive use of reference to indicate correspondences within and between diagrams. They employ an attributed Context-Free Grammar language model to formalize the language model and coordinate recognition during a top-down parse of the input. The grammar is designed so that the parser will recognize objects in decreasing order of reliability, propagating constraints arising from implicit relationships as objects are recognized.

For well-defined documents, syntax-based methods are powerful. However, they can be brittle in the sense that inputs not resulting in a valid interpretation (parse) produce empty input. Returning partial parses and error-correcting parsing [36] may be used to mitigate this problem, but not entirely solve it. Grammars are created by system designers, as grammatical inference remains a hard machine learning problem [37].

While mature commercial document analysis systems now exist, opportunities for improvement remain. Competitions on graphics recognition problems are held regularly, both for lower-level operations such as vectorization and text/graphics separation, recognition of text in diagrams (including rotated and slanted text such as found in the USPTO data; see Figs. 1 and 2), and the interpretation of specific graphic types including technical drawings, tables, flowcharts, chemical diagrams and mathematical notation [20], [19], [21], [38]. These competitions normally consider the recognition of isolated graphics.

In the USPTO competition, all of the top-5 systems use the more common 'generate-and-test/revise' recognition strategy. However, the competition is unusual in that text describing patent diagrams is provided which can be used to improve recognition quality.

### *B. OCR and Text/Graphics Separation*

An important OCR benchmark was the University of Nevada at Las Vegas competitions, held annually during the early 1990's [39], [40]. Since that time, text OCR has become a fairly mature technology, and correspondingly text recognition research has shifted toward the harder problems of recognizing text in natural scenes and videos [41], [42], and documents captured by camera [43]. Over the last decade, there have been a number of Robust Reading competitions on these topics, held as part of the International Conference on Document Analysis and Recognition[5], [6].

A common early processing task is to separate regions containing text vs. other page contents into two or more layers. This process is termed text/graphics separation. Most text/graphic separators filter large connected components in the early stages, along with long/thin and very small connected components. This tends to filter out characters that are small (e.g., '.', ',') or that touch characters or graphics; attempts are made to recover these lost characters during word or text line detection, as described below.

Image features for detecting text have included connected component shape, aspect ratio, density, and spacing [29], [44], similar features for skeletonized connected components [45], and textural features that exploit the relatively high visual frequency of text in comparison to graphics (e.g., using Gabor filters [29] or Gaussian second derivatives [46]). A common challenge is handling different font sizes; this is dealt with by measuring features at different scales [46], [47], [48]. Recently, image patches have been used instead of connected components, along with feature learning methods such as k-SVD and sparse representation [47] and convolutional k-means [48] to construct discriminative image patch feature spaces.

As a more specific instance of text/graphics separation, most of the top-5 systems in the USPTO competition detect tables to avoid confusing text in table cells with part labels, using simple line detection and following. A survey of table detection methods is available [49].

Words and text lines are commonly detected through clustering detected characters. Inter-character, word or text-line distance is estimated by some combination of distance, relative orientation, and similarity of the components (e.g., by estimated text height) in each group. Morphological operations have been used to merge clusters and shrink/tighten boundaries of connected components during clustering [44].

To detect rotated text (e.g., in engineering drawings and maps), Fletcher and Kasturi [50] make use of the Hough transform to determine text line orientations from connected components. Tombre et al. extend their approach using median and linear regression to produce additional local text orientation estimates when clustering detected characters into strings [51]. To focus search, estimated word or text line end points may be used as the initial cluster centers [51], [?]; Roy et al. use the shape of spaces between characters while expanding the ends of text lines, and are able to extract curved text, such as found in document seals [?]. Bukhari et al. have provided a recent survey of current methods for detecting curved and warped text lines [52]. One approach estimates baseline and x-line (the 'middle' line that sits on top of a lower case 'x') locations for individual characters, and then places active contours (snakes) at the top and bottom of connected components which are deformed based on an energy model, after which overlapping snakes are combined [52].

Problems encountered in recognizing annotations in engineering drawings such as object dimensions [53] are similar to the USPTO Challenge. As in DIA generally, the use of language constraints in guiding recognition is critical. For example, detected arrows may be used to detect dimension text more reliably [45], [54].

As described in Section IV, the top-5 systems submitted to the USPTO competition employ simple text/graphics separation methods (e.g., using the size of connected components) along with OCR techniques of varying sophistication. In some of the systems, OCR results are used to revise the initial text/graphics separation. Different methods are used to detect text orientation (portrait vs. landscape) and to detect text at other angles (see Fig. 1).

## III. The USPTO Challenge

This section describes the design of the USPTO algorithm competition including the recognition tasks, reference image data and ground truth creation, evaluation and scoring methods, and the competition outcome. Table II summarizes the main tasks for the USPTO competition and their relationship to the Document Image Analysis tasks described in Section II. As input, systems receive a greyscale patent document image, and an HTML file containing a portion of the text of the patent (patents are, by requirement, greyscale). Systems need to locate figure locations, figure titles, part label locations, and the contents of the part labels. An unusual characteristic of the USPTO competition is the availability of accompanying text to aid in graphics recognition, which is infrequent in the literature.

Consistent with usual practices in programming contests, participants were able to make repeated code submissions to enable testing of solutions and gather feedback about solution quality. Participants were organized in teams of two, and both members of a team were able to submit program solutions.[2] Submissions were compiled and executed on central competition servers where the solutions were tested against a private set of test case images to allow objective scoring. The overall team score was given as the maximum score of both team members.

Three different datasets were created for the competition (see Section III-C for details). A training dataset was available to all participants, which they could use to design and test code on their own machines. During the contest, any system submission by a participant to the contest site would generate a score visible only to the team using a second private dataset (the system test set). To prevent over-fitting of submissions to the training or system test datasets, the last submission of each team was re-scored using a third, private evaluation dataset at the end of the contest.

Solutions submitted to the competition were evaluated using a program that called two functions defined in the submitted program, `getFigures()` and `getPartLabels()`. For system tests and final scoring, the competition imposed a time limit of one minute per test case and a memory limit of 1024 MB. There was no explicit code size limit but a limit of around one MB was advised. Furthermore, the binary executable size was limited to one MB, and the compilation time limit was 30 seconds.

System scores on the final test set were used to rank systems and award prizes. System scoring is described in Section III-D. The execution, prizes, and outcome of the competition are described next.

[2]Embedded within this competition was a social science experiment to investigate different team formation mechanisms. Two treatments were implemented. In treatment one, teams were formed through bilateral agreement between participants after communicating through a public forum or private direct messaging (this was termed the 'free-form' treatment). The second treatment, teams were formed based on a stable-matching algorithm using participants' stated preferences (termed 'algorithm' treatment). We found no significant differences in algorithm performance between the two treatments. The exact details of the social science experiment are beyond the scope of this paper.

### A. Running the Competition

The contest was run on the TopCoder.com online programming competition website, a commercial platform established in 2001 [18], [55], [12]. Working with TopCoder provides convenient access to a standing community of over 400,000 software developers who regularly participate in crowdsourcing competitions, and provides infrastructure for online test and scoring of solutions. (TopCoder also had a working relationship with NASA's Center for Excellence as described above, which allowed the USPTO to pay the cash prizes.) Apart from developing conventional software solutions, competitors on this crowdsourcing platform also regularly compete in contests to solve abstract algorithmic problems that require a mix of logic, mathematical, and programming skills. Since its inception a decade ago, the platform has run over 10,000 competitions.

Given the complexity of the task to be solved, the competition ran for four weeks between the end of 2011 and beginning of 2012 (many TopCoder competitions run only for 10 days).[3] To attract participants, we offered a combined prize pool of $50,000 which was split into two global overall prizes and 22 smaller prizes for virtual competition rooms. We offered two highly attractive global prizes of $10,000 and $5,000 for the first- and second-placed teams. However, offering large, but few overall prizes may not result in the best outcome due to an effort-reducing effect of greater rivalry [56], [55]: if everyone competes against everyone else, an individual team's likelihood of winning may be too low to warrant the investment of substantial effort. Therefore, we organized the competition into 22 virtual rooms, each of which offered an additional $1,000 and $250 'room prize' for the room winner and runner-up. Furthermore, all active participants also received a limited edition t-shirt to acknowledge their efforts in participation which was paid for by the Harvard-NASA Tournament Lab.

Because the overall winning team would automatically take first place in the virtual room, the overall cash prize for the winning solution was $11,000. Within each team, the prize money was split based on peer effort allocation which was determined through a survey conducted after the competition but before winners were announced (in case of disagreement of peer allocation, the mean of both allocations was used). During the competition, no overall ranking or leaderboard was published in order to minimize strategic behavior of participants.

The challenge drew 232 teams (463 participants), of which 70 teams (30%) submitted software code. Twenty-nine countries were represented among those participants who submitted solutions. This group of submitters included 49% professionals, 39% students, and the remainder reporting not working or part time. The bulk of participants ranged between 18 and 44 years old. Seven of the participants were academics (PhD students, professors, or other research positions). Most (80%) of the non-student participants were self-described software

[3]As a side note, we also ran a follow-up competition to build on solutions submitted in the initial competition and further improve detection accuracy. The details of the follow-up competition are outside the scope of the paper.

TABLE II
USPTO Challenge Image Analysis Tasks

| Task | Sub-Tasks |
|---|---|
| Segmentation (Detection) | Text/graphics separation<br>Character regions<br>Part label regions<br>Figure regions* |
| Classification (Labeling) | Character recognition (OCR)<br>Part label text<br>Figure title text |
| Validation | Language model: page/word layout, word syntax<br>Correction mechanisms (e.g., contextual constraints) |
| Coordination (Search) | Hypothesis generation<br>(segmentation, classification and correction)<br>Hypothesis ranking and selection<br>Sequencing hypothesis generation, selection, and validation |
| Model Fitting | Training data; segmentation, classification, validation and coordination parameters |

**Note** :*Figure titles included in figure regions.

developers of various kinds.

Collectively, teams submitted 1,797 solutions that compiled on the competition servers, averaging to 25.7 submissions per team. The submitted solutions used five programming languages (C#, C++, Java, Python, VB). Participants reported spending an average of 63 hours each developing solutions, for a total of 5,591 hours of development time.

The winning team had members from the United States and South Africa, both holding a masters degree and working full time. Their stated professions were 'Computer Vision Scientist' and 'Image Processing Researcher.' Both reported to have much experience in machine learning, pattern recognition, OCR, image processing, and computer vision (mean of 5.7 on scale of '1=No Experience' to '7=Expert Experience'). The winning solution was written in C++. The top scoring algorithm achieved 89.16% precision and 88.98% recall for figure boundary detection, 78.86% precision and 78.76% recall for figure title recognition, and 72.14% precision and 69.86% recall for part labeling.

## B. Image Analysis Tasks and System Design

Designing a system for this competition requires selecting features and algorithms (e.g., template matching, neural networks, agglomerative clustering) for addressing each of the tasks listed in Table II. This leads to the related problem of model fitting, where parameters must be set. Finally, segmenters, classifiers and validators need to execute in some sequence, alternating between generating and selecting hypotheses, which can be termed the coordination of decisions, or *recognition strategy*.

For a given input, the recognition strategy defines a search over possible interpretations of the page image, which are graphs over image regions.[4] Constructing the final interpretation requires a series of segmentation (*where*), classification (*what*) and parsing (*relationship*) decisions, as discussed in Section II. While no parsing tasks are indicated in Table II, relationships between objects must be considered, such as for joining characters into words and correcting characters using surrounding characters (i.e., context).

The language models defined in each system constrain the space of possible interpretations: this includes assumed locations for headers, titles, labels and figures in a page, character and word spacing, and syntax of proper figure titles and part labels. Language models are used to constrain hypothesis generation and validate detection and character recognition results. In some cases detections may be determined to be partially valid and then corrected, such as by replacing '1' by 'i' in the detected string 'F1g.'

## C. Patent Page Image and Text Datasets

For the purposes of this online competition, we prepared a representative corpus of 306 patent drawing pages from various different patent classes. For some patents we included one drawing page in the set and for other patents we included multiple drawing pages. The whole corpus was divided into three subsets A (train), B (system test) and C (final test) containing 178, 35 and 93 drawing pages, respectively. The division was made randomly but with one restriction: all drawing pages belonging to the same patent were always placed into the same subset. We chose this approach to test generalization towards the full patent archive which contains many patents with more than one drawing page. In addition to image data, participants also had access to the patent text ('Claims' and 'Description' sections) in HTML format. For recent patents, the text is relatively error-free, but for older patents the text was automatically OCR'd and so is likely to have errors.

The test dataset was made available in the following format. Folders 'images' and 'texts' contained page images in JPEG format and patent texts in HTML format, correspondingly. For each image 'images/NAME.jpg', the expected correct list of figures and part labels was stored in 'figures/NAME.ans' and 'parts/NAME.ans', correspondingly. Both files were in a similar format. The first line contained an integer – the number of figures (part labels) in the file. The following lines described one figure (part label) in the following format:

$$N \;\; x_1 \;\; y_1 \;\; x_2 \;\; y_2 \ldots x_N \;\; y_N \;\; title(text)$$

where $N$ is the number of coordinate pairs given, $(x_1, y_1) \ldots (x_N, y_N)$ is a polygon defining the figure (part label) location, and title (text) is the figure's title (part label's text). The same format was expected as output from submitted solutions. We also provided an offline tester/visualizer, including Java source code, which allowed participants to check the precise implementation of the scoring calculation.

To create the ground truth reference standard, we used the image annotation tool LabelMe [58].[5] However, since we had to prevent potential leakage of our final scoring images, we did not employ a completely open crowdsourcing approach but rather had a separate instance of LabelMe set up specifically for our purposes. This private instance was then populated with the collected patent drawing pages and used by two administrative contractors paid by the USPTO to mark up the ground truth references. After cross checking the results, the ground truth of one of them was used for the competition.

[4] *Label graphs* [57] have been proposed for uniform representation and comparison of structure in images and other input types (e.g., handwriting).

[5] http://labelme.csail.mit.edu/Release3.0/

### D. Evaluation Metrics and Scoring

The ground truth for figures and part labels may be represented as a list of bounding boxes:

$$(type,\ ((x_{min}, y_{min}), (x_{max}, y_{max})),\ label) \tag{1}$$

where $type$ is figure or part label, the second element provides the top-left and bottom-right coordinates of an axis-aligned bounding box, and $label$ is the textual figure title or part label.

As described above, ground truth figure and part label regions were manually created as polygons of differing numbers of vertices, and associated with the appropriate figure identifier or part label. All participants in the competition used bounding boxes to represent regions (which is conventional in Document Image Analysis). The evaluation metric converts the ground truth polygons to bounding boxes.

A candidate region bounding box $B_C$ matches a ground truth bounding box $B_G$ when the intersection of the two boxes is as large as some percentage $\alpha$ of the larger of the two boxes:

$$area(B_C \cap B_G) \geq \alpha\ max(area(B_C), area(B_G)) \tag{2}$$

where $\alpha_f = 0.8$ for figures, and $\alpha_p = 0.3$ for part labels. Different $\alpha$ values are used for figures and part labels because of the much smaller size of the part labels.

For a figure, the label contains the figure title without the 'Fig.', 'Figure' etc. indication, e.g., label '1a' is the correct label for Figure 1a. Both figure titles and part label strings are normalized before comparison:

- **Figure titles:** All letters are converted to lower case; characters other than a-z, 0-9, (, ), -, ', <, >, . (period), and / are removed from both strings. The resulting strings must be the same for the ground truth and the algorithm's guess in order for a figure title be correct.
- **Part labels:** The same characters are preserved as for figure identifiers. However, there are a number of special cases, for example where two part labels may be indicated together, e.g., '102(103)' or '102,103' indicating parts 102 and 103. Periods/dots must be removed from the end of part labels. Subscripts are indicated using $<$ and $>$ (e.g., $A < 7 >$ for $A_7$); superscripts are represented in-line (e.g., $123^b$ is represented by 123b).

For a figure or part label to be correct, both the region and label must match. Partial credit is given for matching a region correctly but mislabeling it. The match score $match_s$ for partial matches was 25%, and full matches was 100%. To compute the accuracy, $match_s$ scores are added and used in weighted recall for ground truth regions ($R$), precision of output regions ($P$), and their harmonic mean ($F$, the f-measure):

$$R = \frac{\sum match_s}{|Target\ Regions|}\quad P = \frac{\sum match_s}{|Output\ regions|} \tag{3}$$

$$F = \frac{2RP}{R+P} \tag{4}$$

For a given f-measure accuracy $F$ and run-time in seconds $T \leq 60$, the per-file score for figures or part labels was:

$$score = F\ \times\ \left(0.9 + 0.1\left(\frac{1}{max(T,1)}^{0.75}\right)\right) \times\ 1,000,000 \tag{5}$$

Execution time determines 10% of the final score; to give a sense of the effect of speed on the score, at or under one second incurs no penalty, at two seconds roughly a 4% penalty, at five seconds 7%, and at 25 seconds 9.9%. This is noteworthy, because including execution time directly in a scoring metric is uncommon in the document image analysis literature – time and space requirements are considered but within certain bounds accuracy is often the main consideration.

Each input file was scored twice, once for part labels, and once for figures. Systems were scored with 0 points for a test case if they did not produce output, whether due to reaching the one-minute per input time limit, the 1024 MB memory limit, a system crash, or improperly formatted output. The final system score was defined by the sum of all figure and part label test file scores.

## IV. Participant Solutions

In this section we analyze the top-5 systems submitted to the competition, which are summarized in Tables III, IV, and V. We produced this summary by carefully studying the brief system descriptions provided by the participants, along with the system source code. We organize the tables and the analysis around the tasks identified in Table II.

Participant systems were required to implement two functions, one for figures and another for part labels. For the most part, initial image processing was the same for both functions across the systems. The high-level strategy used in the top-5 systems is summarized in Fig. 3

The tables show a number of differences between the systems, and include deviations from the strategy above. Here we will discuss the most significant differences in approach.

Word recognition and page orientation are performed in different orders by the top-5 systems (see third step in Fig. 3). In particular, as can be seen in Table III, the 2nd and 4th place systems perform OCR before clustering connected components (CCs) into 'words' and estimating the page orientation, while the remaining systems estimate orientation and word locations before OCR. The 4th place system did not attempt to identify tables (step 6) but considers three rather than two page orientations; portrait, along with landscape rotated left, and landscape rotated right. The 5th place system makes use of a third category of connected components for figure text, which is normally larger than that for part labels (these are discriminated from graphics and one another by size).

The 2nd place strategy performs OCR iteratively, adapting estimated font parameters to revise character classification and segmentation results (e.g., when character spacing is re-estimated for figure titles).

TABLE III
TEXT DETECTION, CHARACTER RECOGNITION AND IMPLEMENTATION FOR TOP-5 SYSTEMS

| DETECTION | 1st | 2nd | 3rd | 4th | 5th |
|---|---|---|---|---|---|
| CC Classes | (2) Text Graphics | (2) Text Graphics | (2) Text Graphics | (2) Text Graphics | (3) Text FigText Graphics |
| Size filter CCs | ✓ | ✓ | ✓ | ✓ | ✓ |
| Filter margin/ header | ✓ | ✓ | ✓ |  | ✓ |
| Cluster CCs |  |  |  |  |  |
| - After OCR |  | ✓ |  | ✓ |  |
| - Dist. Metric | Ver. overlap + hor. distance | (2 dirs.) Baseline deviation, height ratio, horizontal distance | Ver. overlap + hor. distance | Ver. overlap + hor. distance; Filter short i, l, 1, chars not in 0-9a-zA-Z | Ver. overlap + hor. distance |
| - Table Filter | If ruling lines in 'word' perimeter + intersecting 'words' | if ≥ 2 rows & cols, via hor/ver projection + line detection, | If ruling lines in perimeter and not near an incoming line |  | If lines found at two of four 'word' BB sides |
| Orientation (portrait/ landscape) | Text CCs vote by wider/taller + max hor. vs. vert. overlap with nearest Text CC | OCR and seg. words (both dirs); max mean word width vs. height | Text CCs vote by wider/taller | **Portrait, Landscape L/R**; Consider OCR'd Text CCs in 3 directions | Text CCs vote by wider/taller |
| RECOGNITION |  |  |  |  |  |
| Char. Classes | (14)<br>0-9fgac | (36)<br>()0-9a-hj-np-z | (22)<br>0-9a-dA-F | (31)<br>0-9a-giruA-GIRU!<br>(69)<br>()0-9a-zA-Z-/.,! | (56)<br>()0-9a-lnr-uxy<br>A-TV-Z |
| Features | 16 x 32 ternary (b/w + hole) | 16 x 16 grey floats (mean subtracted); De-slanting; add empty areas for empty ascender/descender regions | W x 40 grey (avg. intensity); W is avg. width for each class | 6 x 6 density (% black), + width, height, aspect ratio, total density | 15 x 15 binary |
| Classifier | Template (% pixels match) | 256:50:36 MLP | Template (sum squared difference of pixel grey values) | max of<br>40:30:31 MLP<br>40:25:69 MLP | 225:225:56 MLP |
| *Notes* | **Only lower case; No i/I** | **Only lower case; No i/I**; Iterative refinement - re-estimates ascent/descent regions, spacing, italic angle | **No g/G**; 0, 1 have two sub-classes; resizes input to match each template width | First MLP for frequent characters |  |
| IMPLEMENTATION | C++ | C++/OpenCV | Java | C#/R(nnet) | C++ |

RECOGNITION STRATEGY FOR TOP-5 SYSTEMS

1) Filter margins/document header
2) Text/graphics separation. Classify connected components (CCs) as *Text* or *Graphics* by size and aspect ratio
3) Word recognition and page orientation detection
   a) Cluster connected components into words
   b) OCR: recognize individual *Text* CCs as characters
   c) Determine page orientation (portrait or landscape)
4) Reject 'words' identified as tables
5) Identify part labels
   a) Validate 'words' using label syntax and reject labels with low character confidences
   b) Return part label BBs with label text
6) Identify figure titles and locations
   a) Validate 'word' character strings using title syntax
   b) Cluster words around detected figure titles
   c) Return figure BBs and titles

Fig. 3. Recognition Strategy Used by the Top-5 Placing Systems. The order of sub-tasks for Step 3 varies between systems

The most distinct figure title and region recognition strategy is that of the 3rd place system. Rather than matching patterns in recognized character strings, a word-shape model based on the geometry of 'Fig', 'Figure' etc. is used to locate possible titles before applying OCR, after which this word shape model is updated to match the best candidate (determined by location of the title and character confidence). Further, figure regions are obtained using a top-down X-Y cutting approach [3], rather than bottom-up agglomerative clustering of words or CCs.

## A. Model Fitting

All systems use thresholds to define minimum and maximum CC sizes to include in the analysis, and thresholds on distances for an initial clustering of CCs into 'words.' A number of the systems make use of thresholds on character recognition confidence in ranking and selecting figure title and part label hypotheses (e.g., in the 1st place system, to control correction of part labels). Additional parameters are indicated in Tables III – V; to see all parameters for each system, the source listings may be consulted.

For the template character classifiers (1st and 3rd place), templates are defined using training images: the 1st place system uses a small set of characters directly taken from training images for just the characters *0-9, f, g, a* and *c*, while the third place system uses approximately 5,000 images, which are then averaged to produce a template for a much larger character set (see Table III).

For the neural network (multi-layer perceptron) classifiers, the 4th and 5th place systems use characters in the training images, while the 2nd place system uses characters from the training images along with synthetic characters generated using fonts with different parameters, for a total of 145,000 training samples (95% of which are used to train the final classifier). Characters from training images are transformed to 8 different slant angles, and 89,000 synthetic characters are created using different variations of the 'Hershey' font, along with variations in thickness and slant.

The 2nd place system is unique in that font metrics used for classification are iteratively re-estimated at run time. Some of the other systems adapt parameters at run time as well but to a lesser degree. One example of this is that a number of the systems discard 'words' whose size differs significantly from the mean size of detected words.

TABLE IV
FIGURE AND PART LABEL DETECTION FOR TOP-5 SYSTEMS

| FIGURES | 1st | 2nd | 3rd | 4th | 5th |
|---|---|---|---|---|---|
| Label Syntax | Contains ‘f1g’ (fig) | Patterns for ‘fig’ (‘f1g’, ‘fzg’, etc.) | **(Word-shape)** FIG + 1-3 digits, FIGURE + 1-2 digits | Starts with ‘fig’ | 3-9 chars; contains ‘fig’/‘FIG’ legal letters: a-d, A-D |
| Label Detection | Match words containing ’f1g’ | Match ‘fig’ patterns; estimate font attributes, repeat OCR + look for ‘fig’ patterns; match words with first three letters with similar ascent, descent, w/h as detected labels; re-estimate spacing for large fonts; repeat OCR on candidates | Match word shape models; find best candidate by location in figure and char confidence, discard candidates highly dissimilar from ‘best’ candidate | Match words starting with ’fig’ | OCR FigText CC cluster candidates, reject illegal labels |
| Regions | Merge words with nearest fig. label; random search to min. penalty to obtain one label per figure | Merge words starting with figure labels; iteratively add CCs, updating BB, allowable growing directions. Obtains k regions for k labels (k=1 if no fig labels detected) | **(Top-down)** Cut page at horizontal or vertical gap until each figure contains one label | Merge CCs by proximity (omitting fig labels); obtain k figure regions for k figure labels; assign labels to regions by minimum total distance | Merge non-Graphics CCs with closest Graphics CC. Detect figure labels, then assign merged CCs to closest figure label |
| PART LABELS | | | | | |
| Label Syntax | Max 4 chrs, at least 1 digit, no more than 1 alphabetic; reject legal Fig. labels | **Max 9 chrs**, up to 2 alphabetic; penalize labels starting with alpha. char; cannot contain ‘mm’; reject ’0’ ’l’ ’1’ and one alphabetic char | Max 4 chrs, up to 2 alphabetic | Max 4 chrs, containing digits, digits followed by letters or letters followed by digits. | Max 4 chrs, cannot be an isolated 0 (zero) or 1 (one) |
| Label Detection | Filter large/small; reject low confidence OCR, illegal labels | Reject illegal; score candidates by char confidences + unlikely label penalties (e.g. ‘a1’); re-estimate font parameters on most confident candidates; reject characters far from mean confidence, OCR modified labels | Reject figure labels; most frequent text height for part labels is estimated as most frequent in Text CCs; reject low confidence labels not near a connecting line | Reject illegal labels | Find small Text CCs; merge adjacent characters; OCR and reject illegal labels; reject if overlapped by fig labels |

Many parameters are set based on assumptions, or ‘empirically’ by trying different parameter values and selecting those that perform best. Practically speaking, some parameters must be set in this manner – for example, it would be impractical to try all possible image sizes to use as input to a classifier (e.g., from 1 pixel up to different aspect ratios for full resolution), and even less practical to consider them with *combinations* of other parameters: the combinatorics get out of hand quickly, as well as the system designer’s ability to compile and analyze the results from parameter combinations.

### B. Validation

An interesting difference between the top-5 systems are the language models used for characters and labels, as seen in Tables III and IV. The 1st place system makes use of only 14 character classes, with no class for i or I, which class ‘1’ is expected to catch. The reason that this does not lead to terrible performance is the correction mechanism used. For both figure titles and part labels, words extracted from the patent HTML text using pattern matching define a set of possible output strings, and hypothesized figure titles are matched to their most similar sequence in the patent text, and low-confidence part-labels are corrected using the most similar part label detected in the HTML text (see Table V). The 1st place system was the only one that used the HTML text to validate and constrain labels. Note that fixing the label outputs to those found in the patent text reduces the output space, as it is likely much smaller than $|C|^n$ possible strings, where $|C|$ is the number of character classes, and $n$ the length a string (for 4 characters taken from 14 classes, this is 38,416 strings before applying syntax constraints).

The language models for figure titles are fairly similar across the systems, with some variation of ‘fig’ assumed to be at the beginning of a figure title. Most systems assumed that part labels are four characters long except for the 2nd place system, where up to nine characters are permitted. All systems reject labels that are inconsistent with their language model in order to avoid false positives, and some reject labels containing low confidence characters.

## V. RESULTS AND DISCUSSION

All submissions were scored and then ranked using the method outlined in Section III-D, using the final test set of drawing pages and HTML texts (93 drawing pages, described in Section III-C). This determined the ranking of systems shown in Tables III - V. Cash prizes were awarded based on the system ranking. We find a high correspondence in ranks between the two tasks (Kendalls tau rank correlation of $0.511$; $p < .05$ for the first ten ranks) indicating that those teams doing well in one task, were also doing well in the other. No system outside the top-5 scored higher on any individual task (figure or part label detection). Consequently, the discussion of the top-5 solutions covers the best approaches that were submitted to the competition.

Fig. 4 shows the variance of algorithm performance for all test cases of the top performing algorithms as an overall score, and decomposed into the scores achieved on the figure detection task and the label detection task. All algorithms fail on at least some test cases in both tasks achieving a score of zero. Conversely, all algorithms also achieve a perfect score for at least some test cases. The figure detection task was significantly easier, with most algorithms receiving a perfect score on many test cases. Thus, performance in the figure detection task was mostly driven by algorithms’ performance on hard test cases.

When considering timing and accuracy results separately, different systems come to the fore as strongest in figure

TABLE V
ADDITIONAL VALIDATION AND CORRECTION FOR TOP-5 SYSTEMS

| VALIDATION | 1st | 2nd | 3rd | 4th | 5th |
|---|---|---|---|---|---|
| Uses HTML | ✓ | | | | |
| Part Labels | For low character confidence labels, assign most similar word found in HTML text to detected word taking common confusions into account (e.g., '5' and 's') - reject if more than 1/2 chars replaced | Separate labels joined by parentheses (e.g., 103(104)') | Single digits must have high confidence | | |
| Figure Labels | Sort extracted figure captions numerically. Match best sequence of figure labels in HTML text using detected page number and header | | Reject labels with low character confidence or unexpected position (i.e., not near boundary of figure region) | | Penalize matching score by not counting unexpected characters |

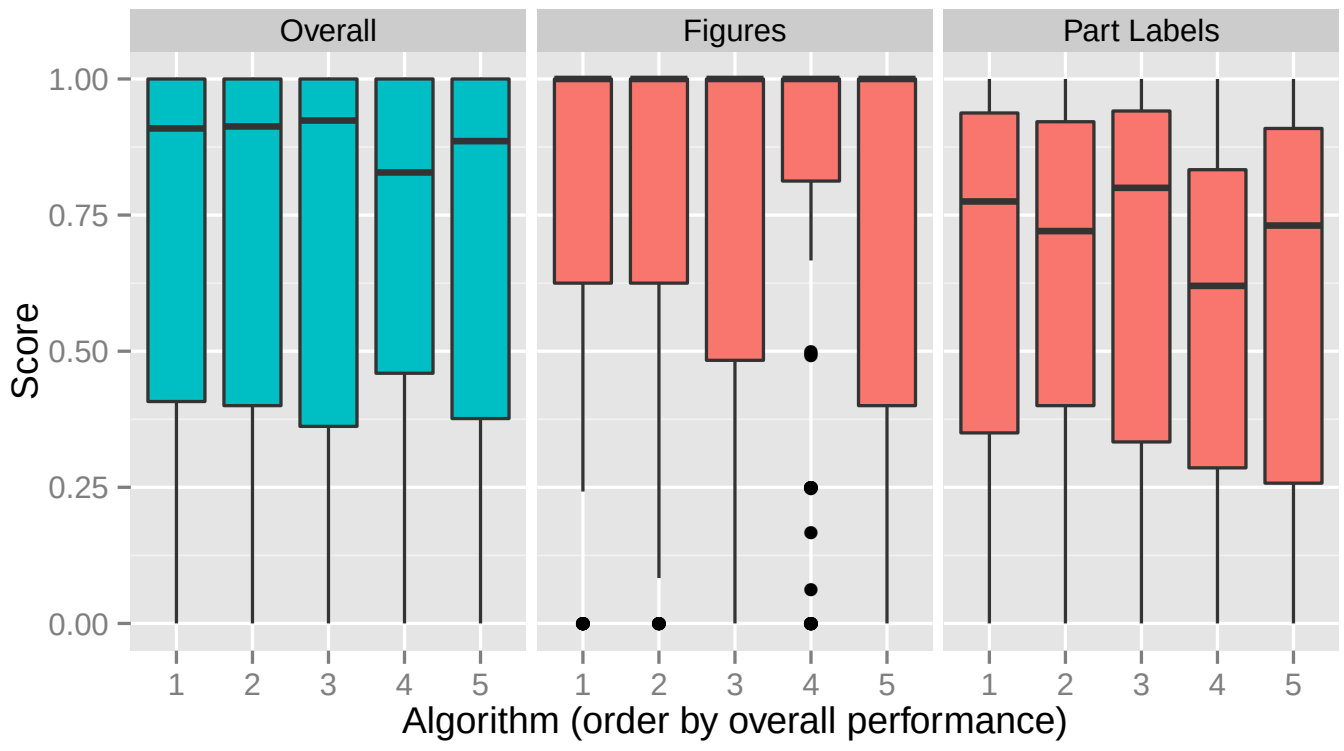

Fig. 4. Boxplot of algorithm performance of individual test case score. At left is overall score (including time taken), at center is weighted F-measure for figure regions and titles, and at right is weighted F-measure for part labels (1.0: perfect score; 0: complete failure).

detection, part label detection, character recognition (OCR), and speed.

### A. Differences in Speed

Average execution speed per test case is shown in Fig. 5. In all cases, part label detection is slower than figure detection and labeling, in part because there are usually many more part labels than figure titles in a patent diagram (one figure is required for there to be at least one part label, but many part labels can be assigned to a figure). All character recognizers used in the top-5 systems should have fast execution since all of them are template or MLP-based. Run times were collected as part of the overall performance evaluation by the TopCoder on their systems. They are measured in milliseconds but it is possible that some of the differences shown arise from operating system interrupts and other confounds.

The top-5 systems fall into roughly two groups according to average speed; the 2nd and 5th place systems are much faster than the 1st, 3rd and 4th-place systems. Some possible explanations are that the 3rd place system resizes character images to fit the different template widths for each class in its character classifier, while the 4th place system runs OCR and analysis in three different page orientations, and runs two different classifiers on every character. The 1st place system uses an iterated random walk to locate figure regions, and its correction mechanisms using the HTML text performs a linear search over all candidate words from the document. While the 2nd system uses an iterated adaptation mechanism, its adaptation is constrained based on character confidences from the classifier. The 5th system is the simplest design of the top-5 systems, using simple geometric and visual features, running OCR just once with simple correction (see Table V).

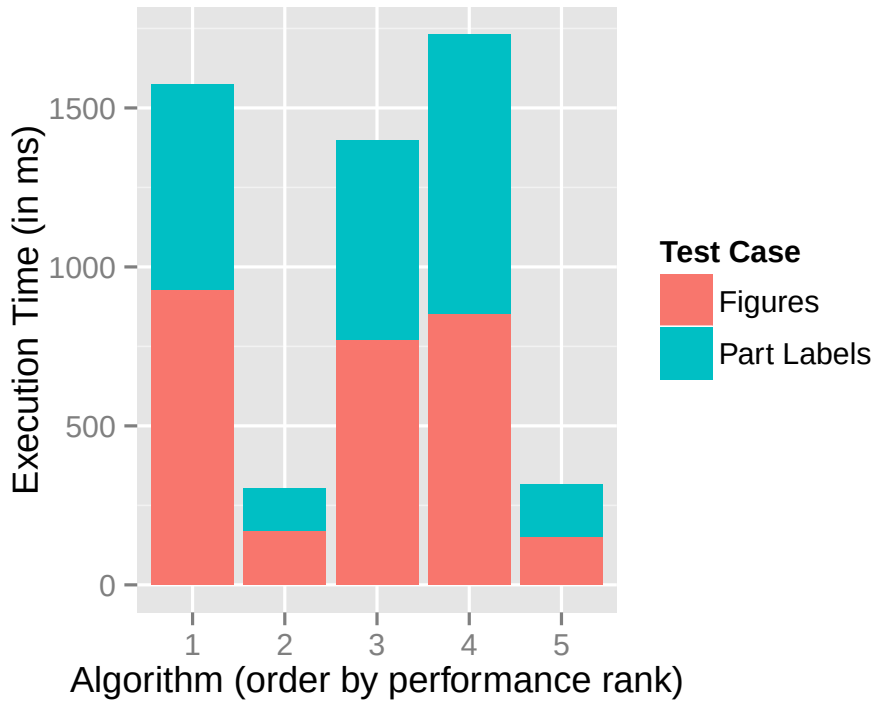

Fig. 5. Average execution time per test case (time given in milliseconds).

### B. Differences in Accuracy

Next, we consider the accuracy of figure and part label region detection (Fig. 6 top panel), and correct region detection and labeling (Fig. 6 bottom panel). In each panel, commonly the best result is taken to be that with the highest F-measure ($2RP/(R+P)$, where $R$ is weighed recall and $P$ is weighted precision - see Section III-D). For a given sum of recall and precision values, the highest F-measure lies along the diagonal from the origin to (1.0,1.0) in the plot. F-measure penalizes both low recall (i.e., many false negatives) and low precision (i.e., many false positives).

The best system for figure detection and labeling is the 4th place system with better than 80% recall and precision (see Fig. 6b) but interestingly was not the best at localizing figure regions. In fact, 4th place performs worst at locating figure BBs amongst the top-5 systems, (Fig. 6a), but the difference between them is relatively small (roughly 5-10%). This system employs two different MLPs for character recognition, and detects figure titles simply by matching 'fig' at the beginning of a segmented word. For the top-5 systems, the region in the R-P plots simply translates by roughly 10%, reflecting the increased difficultly in accurately obtaining figure titles in addition to figure regions. The 1st and 2nd place systems have similar F-measures, and are closest to the 4th place system.

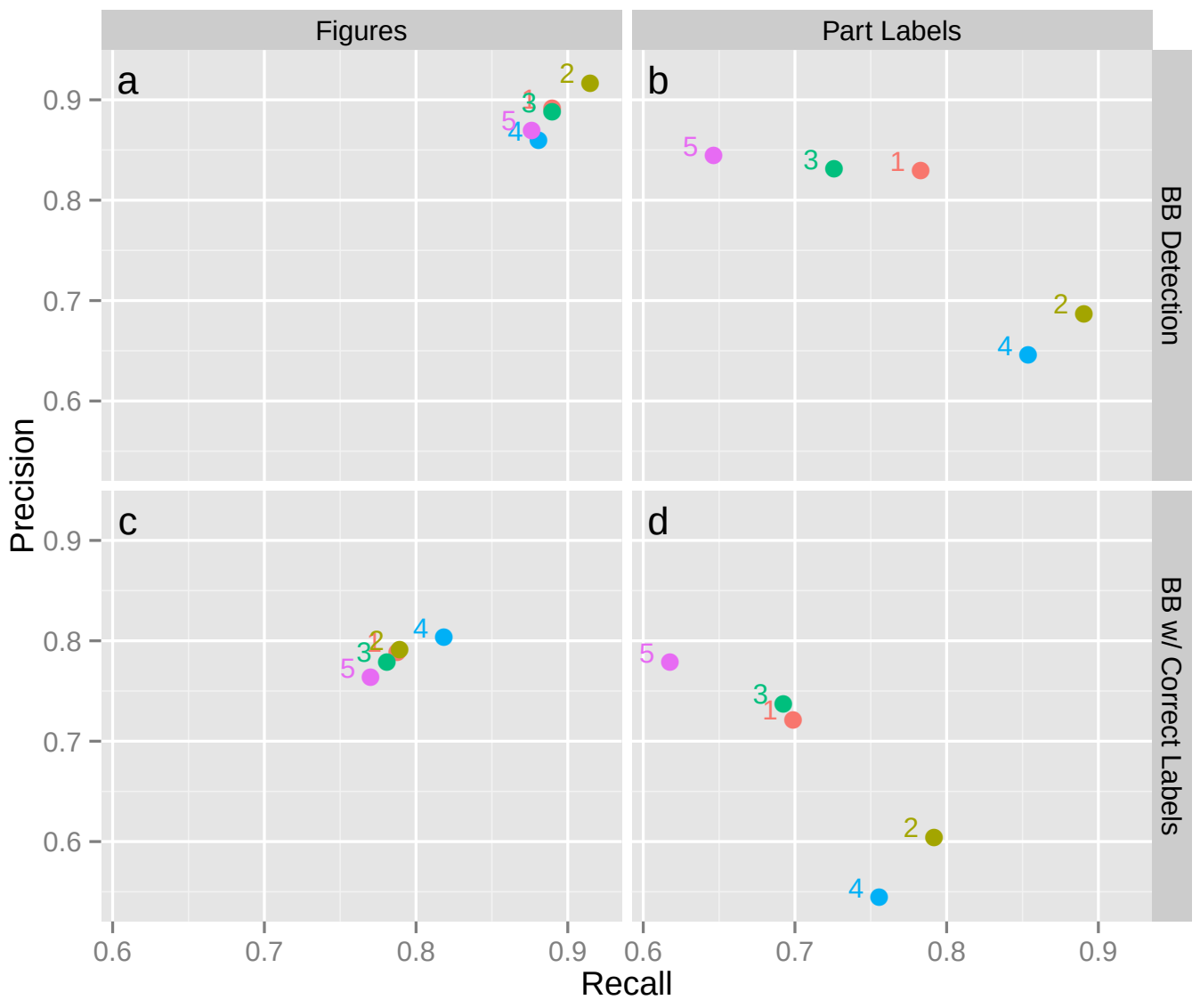

Fig. 6. Accuracy: Recall and Precision Plots for BB detection and BB detection with correct labels.

For part labels the best systems are 1st and 3rd place; once location and label text is taken into account (Fig. 6), the 3rd system has slightly higher precision, and the 1st system slightly higher recall. The 4th and 2nd place systems have roughly 5-10% higher recall but also 10-15% lower precision than the 1st and 3rd systems. The relative performance of the top-5 systems is similar for the BB detection and BB + label metrics plots. Both the 2nd and 4th-place systems apply OCR in multiple directions (2 for 2nd, and 3 for 4th), while the other systems estimate the orientation and then apply OCR in one direction. 2nd and 4th place also have the most sophisticated character recognizers (see Table III). 4th place has the lowest precision, perhaps in part because it does not attempt to detect and remove tables during part label recognition. The high precision but low recall of 5th place might be explained by a well-trained MLP classifier paired with strict rejection of illegal part label strings.

### *C. Comparison System for Part Labels*

After the competition, our team implemented a different system for part labels detection [48] using Python and the Tesseract OCR engine [59]. The results we present below are for an improved version of this system over the previously published version. The improvements include: 1) modifying training label regions to more tightly crop around figure and part label text, 2) CC size filters and header-cutting, 3) using projection profile cutting (using gaps in pixels projected on the x or y-axis) to detect and separate multiple text lines in detected word regions, and 4) refining the part label language, validation and correction approaches.

For the part label language model, character classes were 0-9a-zA-Z; the label syntax allows a maximum number of 4 characters with at most 2 alphabetic, at least one digit, and rejecting isolated 'i' 'I' and '0' characters, and words containing 'Figure,' 'Fig,' etc; letters 'O' and 'o' are replaced by '0.' Like the 1st place system, a set of possible part labels are extracted from the text. OCR is performed twice, using progressively larger paddings around a detected label; if the results differ, whichever has minimum Levenshtein (i.e., string edit) distance [60] with a word in the HTML set is selected.

The system initially detects part labels at the pixel level, using a boosted ensemble of three pixel-level text/non-text AdaBoost classifiers using image patches at different resolutions. Three template dictionaries are learned automatically using a convolutional k-means method for text, background, and both types of pixels in ground truth. Text pixels within a given distance are clustered into candidate words, and then OCR'd and corrected. The final part-label location and label precision and recall measures are 71.91% and 73.55% (F: 72.72%), versus 72.14% precision and 69.87% recall (F: 70.99%) for the 1st place system. The average execution time of the system was 7.52 seconds (measured on a 2-way six-core Intel Xeon X5670 with 96 GB of RAM), with roughly 1-2 seconds for Tesseract, which was run as a separate application on detected regions. The relatively slow speed is explained in part by using Python rather than a faster language such as C, and the time taken to compute the pixel-level features used in the system. In comparison, all top-5 systems have average execution times of under one two seconds (measured on a quad-core Intel Xeon 3.60 GHz with 4 GB of RAM), in part to avoid penalization for slow execution in the scoring metric.

Our comparison system provides a slight improvement in accuracy over the top-5 systems but runs more slowly. We also acknowledge that our system benefits from being designed after the competition results had been published. We feel that this result confirms the quality of the top performing systems submitted to the USPTO competition, given they had significant time and space constraints, and no benefit of hindsight.

## VI. Conclusion

In this paper, we present the results of a month-long algorithm competition to solve a difficult image processing task for the USPTO. In summary, we show in detail the results of using a prize-based contest to recruit skilled software developers to productively apply their knowledge to a relevant image processing task in a practical setting. The resulting diversity in the submitted solutions has the potential to further improve the solution, for example, by implementing a voting mechanism across algorithms. In the simplest case, performance could be improved by combining the best solution for the figure detection task (4th place system) with the best solution for the label detection task (1st place system). The comparison against the performance of a leading alternative implementation confirms the quality of the top performing systems.

The emergence of commercial online contest platforms such as TopCoder, InnoCentive, and Kaggle, which offer access to large pools of skilled software developers, has the potential to enable organizations to crowdsource solutions to difficult software and algorithm development tasks which could not have been developed in-house. This approach could be especially useful to address the demand for the many different and highly specialized image processing algorithms which are required as a result of an explosion available imaging data.

The top algorithms presented in this paper used a variety of approaches and were both fast and accurate. Although the systems perform well, they are not yet accurate enough to be put into every day use. However, the scoring mechanisms that reflect real-world performance considerations, the data, and the top performing solutions are openly available, and we hope that this will stimulate additional research in patent document analysis. Releasing the source code of the five winning solutions makes a breadth of alternative approaches available and offers the opportunity to study the specific causes of differences in performance. The analyses we presented in this work are a first step in that direction. The results of the winning teams provide a great starting point for future developments and the implementation of meta-algorithms leveraging the diversity of submitted solutions, which will hopefully lead to more accurate solutions in future.